\pdfoutput=1
\documentclass{article} % For LaTeX2e
\usepackage{iclr2027_conference,times}

\usepackage{amsmath,amsfonts,bm}

\newcommand{\captiona}{{\em (a)}}
\newcommand{\captionb}{{\em (b)}}

\def\eqref#1{equation~\ref{#1}}
\def\1{\bm{1}}

\DeclareMathAlphabet{\mathsfit}{\encodingdefault}{\sfdefault}{m}{sl}
\SetMathAlphabet{\mathsfit}{bold}{\encodingdefault}{\sfdefault}{bx}{n}

\usepackage{hyperref}
\usepackage{url}
\usepackage{graphicx}
\usepackage{booktabs}
\usepackage{caption} % for \captionof in combined table+figure floats
\usepackage{microtype}
\usepackage{enumitem}

\usepackage{wrapfig}
\usepackage{float}

\newcommand{\erel}{\varepsilon_{\mathrm{rel}}}
\newcommand{\deff}{d_{\mathrm{eff}}}

\title{Apparent Compression, Real Stability: \\ The Intrinsic Dimension of Learning a Quantum Wavefunction}

\iclrfinalcopy
\author{Lu Wei$^{1}$\quad Yufeng Wang$^{2}$\quad Chenfeng Cao$^{3}$\quad Haibin Ling$^{4}$ \\
$^{1}$Data Science Department, $^{2}$Department of Computer Science, Stony Brook University \\
$^{3}$Dahlem Center for Complex Quantum Systems, Freie Universit\"at Berlin\quad $^{4}$Westlake University \\
\texttt{luw744895@gmail.com}, \texttt{yufeng.wang.2@stonybrook.edu}, \\
\texttt{chenfeng.cao@fu-berlin.de}, \texttt{linghaibin@westlake.edu.cn}
}
\begin{document}

\maketitle
\lhead{Preprint. Under review.}

\begin{abstract}
How many directions in weight space does training need? The intrinsic dimension answers this with the smallest number of random directions in which training still reaches a target accuracy, and small values have motivated parameter-efficient methods such as LoRA. We measure it for variational Monte Carlo (VMC), which trains a neural network to represent the ground state of a quantum many-body system. VMC is a demanding test, because the network generates its own training samples and every gradient is noisy, and a revealing one, because the exact answer is known and every run can be scored. We train only a small latent vector that a frozen random map turns into the network's weights, with no change to the standard natural-gradient optimizer. We find that a small dimension can be misleading, while the stability it brings is real. On a magnet with a hard sign pattern, a network that cannot represent signs reaches its best energy in 8 of 28,642 directions, but only because no such network can go lower; once signs are learnable, neither the signs nor the magnitudes are cheap. The dimension rises across a quantum phase transition, so it tracks how difficult a state is at far less compute than fitting a scaling law, yet it never falls below a floor set by the random subspace itself, even where the ground state is nearly trivial. Training in the subspace, in contrast, never diverged in our experiments, whereas full-parameter training with the same settings did, and a control with matched solvers attributes the difference to the reduced dimension.
\end{abstract}

\section{Introduction}
\label{sec:intro}

\looseness=-1 How many directions in weight space does training need? \citet{li2018intrinsic} answered this question by training a network only inside a random low-dimensional subspace of its weights and recording the smallest subspace dimension at which training still reaches 90\% of the full model's performance. They called this number the \emph{intrinsic dimension} of the objective landscape, and measured it for supervised and reinforcement learning. Later work measured it for the fine-tuning of language models \citep{aghajanyan2021intrinsic}, where it motivates parameter-efficient methods such as LoRA \citep{hu2022lora}.

\looseness=-1 This paper brings the measurement to variational Monte Carlo (VMC), the standard way to train a neural network to represent the ground state of a quantum many-body system \citep{carleo2017solving}. Such a network, called a neural quantum state (NQS), assigns to every configuration of the system an amplitude with a magnitude and a sign, and VMC adjusts its weights to lower the energy of the state it represents. The setting differs from the ones above in three ways. First, there is no dataset. The energy and its gradient are estimated from configurations sampled from the network itself, so every gradient is noisy and the noise depends on the current weights. Second, training uses a natural-gradient method called stochastic reconfiguration \citep[SR;][]{sorella1998green,sorella2001generalized}, rather than plain gradient descent. Third, for the systems we study the exact ground-state energy is known, so the error of every run can be computed exactly, which benchmarks in vision, language and control do not offer. To our knowledge no published work measures intrinsic dimension in this setting. It is not obvious in advance that training in a random subspace survives the sampling noise, or what the measured number would mean.

\begin{figure}[!t]
\centering
\includegraphics[width=\linewidth]{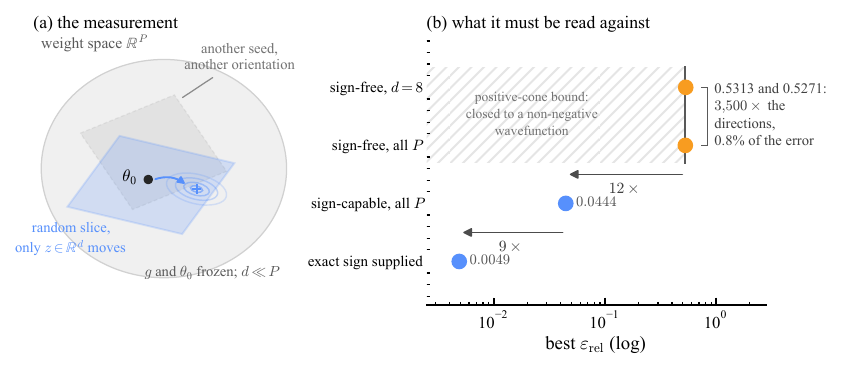}
\caption{\textbf{Method overview.} \captiona\ The measurement. Training is confined to a random $d$-dimensional slice of weight space through the initialization $\theta_0$: the map $g$ and $\theta_0$ are frozen, so only the latent $z \in \mathbb{R}^d$ moves and the slice is fixed before the first step. Its orientation is drawn at random, and a different seed draws a different slice, which is why the seed selects the basin a run ends in (Sec.~\ref{sec:fourarm}). \captionb\ What a measured dimension has to be read against, on the $4\times4$ frustrated magnet. A network with non-negative amplitudes reaches $\erel = 0.5313$ in 8 of its 28,642 directions and 0.5271 with every direction trainable, so $3{,}500\times$ the directions buy 0.8\% of the error: both sit at the positive-cone bound, the lowest energy any non-negative wavefunction can reach. Letting the model represent signs is worth $12\times$, and supplying the exact sign a further $9\times$ (Secs.~\ref{sec:fourarm}, \ref{sec:sign}).}
\label{fig:overview}
\end{figure}

\looseness=-1 We restrict VMC training to a random subspace as follows (Fig.~\ref{fig:overview}). A map $g$ is drawn at random and then frozen. It generates all $P$ weights of the network from a latent vector $z$ of dimension $d \ll P$, and only $z$ is trained, so training moves within a random $d$-dimensional slice of weight space. SR applies to $z$ with no change to the training loop, because the natural-gradient metric of $z$ follows from that of the weights by the chain rule (Sec.~\ref{sec:geometry}). We then define $d^*$ as the smallest $d$ at which the median relative energy error over random seeds falls below a threshold $\tau$ within a fixed training budget. It is the counterpart, at a fixed budget, of the intrinsic dimension of \citep{li2018intrinsic}.

\looseness=-1 We measure $d^*$ on two spin models. The first is a frustrated Heisenberg magnet, where the pattern of signs of the ground state is widely regarded as the hard part to learn \citep{westerhout2020generalization,schurov2025learning}. The second is the transverse-field Ising chain, whose ground state has no signs and which undergoes a quantum phase transition. Besides the linear map of \citet{li2018intrinsic} we test a nonlinear map modelled on Mapping Networks \citep{sen2026mapping}. It trails the linear map on the frustrated magnet and is comparable on the Ising chain (Secs.~\ref{sec:fourarm}, \ref{sec:tfim}).

We propose no new training algorithm. Instead, our main contributions are a measurement and three findings that it yields.
\begin{itemize}
    \item \textbf{Intrinsic-dimension measurement for VMC.} We restrict the training of a neural quantum state to a random subspace and show that SR applies without modification, so the measurement can be added to an existing VMC code as is. Every run is scored against an exact ground-state energy (Sec.~\ref{sec:method}).
    \item \textbf{Subspace training is robust under gradient noise.} In all 246 subspace runs of our sign-structure study, at two lattice sizes, training never diverged, while full-parameter training with the same optimizer settings diverged 41 times in 35 runs at the larger size. A control that reruns the subspace with the full-parameter solver attributes the gap to the dimension. The subspace removes divergence. It does not remove the dependence of the final energy on the random seed (Secs.~\ref{sec:fourarm}, \ref{sec:breakage}).
    \item \textbf{Apparent compression can be a variational bound.} On the frustrated magnet, a network with non-negative amplitudes reaches its best energy in 8 of 28,642 random directions. That energy is the lowest that any non-negative wavefunction can reach, so the small $d^*$ is a property of the model class and says nothing about the ground state. In a model that can represent signs, we give the sign and the amplitude parameters separate budgets and compare with a control that has the exact sign built in. Neither part is cheap. The error keeps falling as amplitude dimensions are added, up to the largest subspace we train, and once the amplitude is resolved the sign needs its whole block of parameters. With every parameter trainable, a learned sign still costs $9\times$ in error against an exact one at equal steps (Sec.~\ref{sec:sign}). The same protocol applies to any method that divides a budget of trainable dimensions among parameter blocks, such as low-rank adaptation (Sec.~\ref{sec:peft}).
    \item \textbf{The measured dimension tracks a phase transition, above a floor set by the random slice.} On the Ising chain, $d^*$ rises about $4\times$ across the transition at two system sizes, at a threshold for which we verify that $d^*$ does not change when the training budget is tripled. Deep in the paramagnet $d^*$ falls only to 64, although the ground state there is nearly a product state. Part of $d^*$ is therefore a cost of the random slice itself (Sec.~\ref{sec:tfim}).
\end{itemize}

\section{Related work}
\label{sec:related}

\noindent\textbf{Intrinsic dimension and parameter-efficient training.}
\citet{li2018intrinsic} measured intrinsic dimension on supervised vision tasks and, with DQN and evolution strategies, on control and Atari tasks. \citet{aghajanyan2021intrinsic} measured it for the fine-tuning of pretrained language models, where 200 parameters projected back into the full space bring RoBERTa to 90\% of full fine-tuning performance on MRPC. LoRA \citep{hu2022lora} builds a method on this low dimension. It freezes the pretrained weights, trains a low-rank update in each layer, and needs 10,000 times fewer trainable parameters than full fine-tuning of GPT-3 175B. For supervised networks, the threshold dimension of a random affine subspace rises as the target loss falls and falls as the initial loss falls \citep{larsen2022degrees}. Our $d^*$ is likewise defined at a fixed threshold $\tau$ and a fixed initialization $\theta_0$. The lottery-ticket hypothesis \citep{frankle2019lottery} makes a related claim in a different basis, a small sub-network found by search, in place of a small random subspace that involves no search. Our work returns to the measurement. In VMC the objective can only be sampled, as in the control and Atari tasks, but exact ground truth allows two controls that those tasks do not offer. An exact-sign control separates the amplitude from the sign, and a product-state limit exposes the floor set by the random slice.

\vspace{1mm}\noindent\textbf{Weight generation through fixed maps.}
Mapping Networks \citep{sen2026mapping} are a nonlinear counterpart of the linear subspace. They generate all weights from a latent vector through a frozen orthogonal map that the latent modulates, followed by a pointwise nonlinearity, and they train with auxiliary stability, smoothness and alignment losses. In quantum tomography, HyperRBMs instead let a hypernetwork modulate the biases of a restricted Boltzmann machine as a function of a Hamiltonian parameter \citep{tonner2026parametric}. Neither work measures the smallest trainable dimension, which is what we study.

\vspace{1mm}\noindent\textbf{NQS optimization.}
Neural quantum states were introduced by \citet{carleo2017solving}. Our base ansatz is an autoregressive NQS \citep{sharir2020deep}. VMC training is preconditioned by SR \citep{sorella1998green,sorella2001generalized}, whose metric is the quantum geometric tensor \citep{stokes2020quantum,vicentini2022netket}, and minSR \citep{chen2024empowering} and an exact sample-space form of SR \citep{rende2024simple} make it affordable at large parameter counts. Parameter-efficient training is also used for NQS. Fine-tuning only the output layer transfers a pretrained NQS across couplings \citep{rende2024finetuning}. Closest to our study, \citet{rende2026scaling} fit compute scaling laws for NQS and assign each Hamiltonian a compute exponent, at a cost of ${\sim}10^5$ A100 GPU-hours.

\vspace{1mm}\noindent\textbf{Sign structure as the hard part.}
Recent work points to the sign, or phase, of the wavefunction as the hard part from two directions. In complex-valued NQS the phase-gradient estimator can be the bottleneck of training, which low-variance estimators relieve \citep{xue2026lowvariance}. The learning complexity of many-body sign structures has been analyzed with Boolean Fourier analysis on exact ground states \citep{schurov2025learning}. Given the exact signs of the same model at $6\times6$, \citet{bukov2021learning} still did not reach a more accurate state. Our dissection measures the same cost inside VMC training, both in dimensions and in optimization. \citet{doschl2025towards} study the interpretability of NQS, whereas we study their trainable dimension.

\vspace{1mm}\noindent\textbf{Machine-learning probes of phase transitions.}
The intrinsic dimension of the data, measured on Monte Carlo spin configurations, has a \emph{minimum} with universal finite-size scaling at continuous transitions \citep{mendessantos2021unsupervised}. We measure the intrinsic dimension of training instead, and find a rise across the transition that relaxes deep in the paramagnet to a floor set by the random slice (Sec.~\ref{sec:tfim}). The two quantities behave in opposite ways at the transition, so they describe different geometries. Weight-space analyses detect phase transitions in the weights of \emph{trained} NQS \citep{hernandes2025adiabatic}. $d^*$ constrains the training itself, so the two approaches are complementary.

\section{Method: Subspace-Restricted VMC}
\label{sec:method} 

Our method, \textit{i.e.}, subspace-restricted VMC, offers three reusable pieces: a measurement of intrinsic dimension for an objective that can only be sampled (Sec.~\ref{sec:geometry}), a budget protocol that tells a block with too few dimensions from one that is hard to learn (Sec.~\ref{sec:sign}), and a health guard for latent training (Sec.~\ref{sec:health}).

\subsection{Setup and notation}
\label{sec:setup}

\looseness=-1 Let $\theta \in \mathbb{R}^P$ be the full weight vector of a base NQS ansatz with $P$ parameters, and $\psi_\theta(\sigma)$ the (unnormalized) wavefunction amplitude it assigns to spin configuration $\sigma$. Training minimizes the variational energy $E = \langle\psi_\theta|H|\psi_\theta\rangle / \langle\psi_\theta|\psi_\theta\rangle$ of a Hamiltonian $H$, estimated by Monte Carlo sampling from $|\psi_\theta|^2$ with error-of-mean $\sigma_E$. With $E_0$ the exact ground-state energy, accuracy is the relative error $\erel = (E - E_0)/|E_0|$. We write $\varepsilon$ for $\erel$ where unambiguous. Subspace restriction trains only a latent $z \in \mathbb{R}^d$, $d \ll P$, through a frozen map $g$ with $\theta = g(z)$ (Fig.~\ref{fig:overview}a). %
We write $\lambda$ for the generic Hamiltonian coupling being scanned.

The base ansatz is an autoregressive neural network (ARNN) wavefunction $\psi_\theta$ \citep{sharir2020deep}. Normalization and lattice symmetry are built into its architecture, so a frozen weight map cannot break them and no constraint on the weights is needed. We study two systems.
\begin{itemize}
    \item \textbf{Frustrated testbed.} The 2D $J_1$--$J_2$ Heisenberg model has nearest-neighbor coupling $J_1$ and next-nearest-neighbor coupling $J_2$, and its frustration is tuned by $J_2/J_1$. We use a $4\times4$ periodic lattice at $J_2/J_1 = 0.5$, where exact diagonalization gives $E_0 = -8.4579233514$ and the network has $P = 28{,}642$ parameters. A $6\times6$ extension appears in Secs.~\ref{sec:sign} and \ref{sec:breakage}.
    \item \textbf{Sign-free testbed.} The transverse-field Ising model (TFIM) is an Ising spin chain with coupling $J$ and transverse magnetic field $h$, and it has a quantum phase transition at $h/J = 1$. We use chain lengths $L = 16$ and $L = 24$. In Sec.~\ref{sec:geometry} the symbol $J$ denotes a Jacobian, and the TFIM coupling appears only in the ratio $h/J$.
\end{itemize}

\subsection{Four training configurations}
\label{sec:arms}

We compare four training configurations, named TC1 to TC4, as summarized in Table~\ref{tab:config}. All four start from the same initialization $\theta_0$ of the same ansatz and use the same training loop and sampler. They differ in which coordinates are trainable. TC1 trains all $P$ parameters and is the reference. TC2 trains $d$ coordinates of a random linear subspace through $\theta_0$, a per-group variant of the construction of \citet{li2018intrinsic}, and is the primary instrument. TC3 trains $d$ coordinates through the same map followed by a tanh, our variant of the Mapping Networks construction. TC4 trains only the last parameter group and freezes the rest, a control that asks whether a fixed, cheap restriction already suffices. Comparisons are made at equal step counts, and between TC2 and TC3 at matched $d$. %

\begin{center}
    \captionof{table}{Four training configurations in our study.}
    \label{tab:config}
\setlength{\tabcolsep}{4pt}
\footnotesize
\begin{tabular}{llll}
\toprule
Config. & Construction & Trainable dim. & Role \\
\midrule
TC1 & full $\theta$ & $P$ & full-parameter reference (minSR) \\
TC2 & $\theta_k = \theta_{0,k} + s_k\, W_k z_k$ & $d$ & linear subspace (Li et al.) \\
TC3 & $\theta_k = \theta_{0,k} + s_k \tanh(W_k z_k)$ & $d$ & nonlinear mapping (tanh variant) \\
TC4 & last parameter group $:= z$, rest frozen at $\theta_0$ & last-group size & cheap-subspace control \\
\bottomrule
\end{tabular}
\end{center}

\looseness=-1 The construction shared by TC2 and TC3 works per group, over the ansatz's top-level parameter groups indexed by $k$. Here $\theta_0$ is the standard initialization, with $\theta_{0,k}$ its group-$k$ block. The matrix $W_k$ is frozen and has orthonormal columns, from the QR decomposition of a seeded Gaussian. The block $z_k$ is the part of the latent allocated to group $k$. The scale $s_k = \mathrm{std}(\theta_{0,k})$ keeps generated perturbations at that group's natural initialization scale.

\looseness=-1 TC3 differs from the published Mapping Networks construction \citep{sen2026mapping}. We omit their latent-driven weight modulation and their auxiliary stability/smoothness/alignment losses, training the latent against the energy alone. In their image-classification ablation the unmodulated variant trails the full method by 2--4\%, so TC3, which omits the auxiliary losses as well, likely understates what the full construction could reach. The latent budget $d$ is split across parameter groups in proportion to group size, with a minimum of 1 and a cap at the group size. We write $\deff$ for the realized total. The effective map seed is the map seed plus the run seed, so different seeds also randomize the frozen map, as in the repeated runs of \citep{li2018intrinsic}.

\subsection{Stochastic reconfiguration in the latent space}
\label{sec:geometry}

\looseness=-1 SR needs the quantum geometric tensor (QGT) $S_\theta$, the Gram matrix of the log-derivatives $\partial_\theta \log\psi_\theta$. Because $z$ is the only trainable collection, automatic differentiation yields log-derivatives directly in $z$-space, and SR operates with the pullback
\begin{equation}
S_z = J^\dagger S_\theta J, \qquad J = \frac{\partial \theta}{\partial z},
\end{equation}
where $J$ is the Jacobian of the frozen map. The training driver, the VMC loop of NetKet \citep{carleo2019netket,vicentini2022netket}, is unchanged. The natural-gradient geometry of VMC is therefore automatically the correct geometry of the latent space. The implementation is uniform across configurations. We use plain SGD under SR, with no adaptive optimizer stacked on the preconditioner. SR uses a dense QGT, an explicit $d\times d$ tensor, for every trainable dimension $\le 4096$, that is, for TC2 to TC4, and minSR for TC1. We control for that split by rerunning TC2 at $6\times6$ and $d = 3072$ on minSR, with 21 runs over all seven couplings at a median of 1120 steps. These runs also log zero divergence events, so the stability contrast survives solver matching.

\subsection{Health protocol for noisy-objective training}
\label{sec:health}

\looseness=-1 Divergence events are counted, logged, and reported as part of the result. An event is declared when the Monte Carlo energy estimate is not finite, jumps upward, or violates the \textbf{variational floor}, that is, falls below the exact energy $E_0$ by more than the sampling noise allows. On an event we roll back to the best variables, halve the learning rate and rebuild the driver, and we abort after a maximum number of resets.

\looseness=-1 The floor guard is necessary when the objective is noisy. Detectors that only catch NaNs and upward jumps, which is the standard pattern in the field, miss corruption of the estimator in the downward direction, which a frozen map can amplify. Health-aware monitoring of training trajectories is also emerging in NQS practice \citep{wang2026nqsagent}. The thresholds, and an incident that only the floor guard catches, are in Appendix~\ref{app:floor}.

\subsection{The observable}
\label{sec:observable}

The observable is the smallest $d$ at which the median error over seeds meets a threshold:
\begin{equation}
d^*(\lambda;\tau) := \min\{\, d \ :\ \mathrm{median}_{\text{seeds}}\ \erel(d,\lambda) \le \tau \,\}
\end{equation}
\looseness=-1 at coupling $\lambda$ and accuracy threshold $\tau$, under a \emph{fixed} budget of steps or compute per run (Fig.~\ref{fig:overview}b). %
The median over seeds is essential, because the orientation of the random subspace selects among distinct attractors (Sec.~\ref{sec:fourarm}), so values from a single seed are not stable. As defined, $d^*$ depends on the budget. Sec.~\ref{sec:tfim} shows at which threshold it becomes independent of the budget in practice.

\section{Experiments}
\label{sec:experiments}

\subsection{Subspace training survives gradient noise}
\label{sec:fourarm}

\vspace{1mm}\noindent\textbf{Setup.} \looseness=-1 We run the $4\times4$ $J_1$--$J_2$ model at the maximally frustrated coupling $J_2/J_1 = 0.5$, on the latent grid $d \in \{8, 32, 128, 512\}$, with 3 seeds per cell. Every configuration is compared within the first 880 steps, the number TC1 completed.

\vspace{1mm}\noindent\textbf{Stability under gradient noise.} \looseness=-1 All 9 runs of TC3 at $d \ge 32$ finished with no divergence event and no abort, and there were \textbf{zero divergence events across all 24 latent-subspace (TC2/TC3) runs}. Latent-only optimization under intrinsic gradient noise is viable, and Sec.~\ref{sec:breakage} shows it is \emph{more} stable than full-parameter training at scale.

\vspace{1mm}\noindent\textbf{A shared bound.} \looseness=-1 Every high-capacity configuration converges to the same $\erel \approx 0.5271$ (Table~\ref{tab:fourarm}): full TC1, last-layer TC4 ($\deff = 578$), and TC2 with best seeds down to $d = 8$. That is $E = -3.99975$, within $0.006\%$ of the smallest diagonal element of $H$, $-4$, attained by the classical N\'eel and stripe configurations. This value does not depend on the network. Every off-diagonal element of $H$ in this basis is non-negative, so no wavefunction with non-negative amplitudes can go below that diagonal minimum, and any amplitude function concentrated on those configurations attains it. We call it the positive-cone bound. Reaching it in $d = 8$ of $P = 28{,}642$ dimensions, an apparent $3{,}500\times$ compression, therefore says nothing about the amplitudes of the ground state. Section~\ref{sec:sign} removes the bound with an exact-sign control.

\begin{wraptable}{r}{.42\linewidth}
    \centering
    \setlength{\tabcolsep}{4pt}
    \captionof{table}{Each row gives the minimum and median best $\erel$ over seeds for one configuration at one $d$, within the first 880 steps.}
    \label{tab:fourarm}
    \footnotesize
    \begin{tabular}{lrrr}
    \toprule
    Configuration & $\deff$ & min $\varepsilon$ & med. $\varepsilon$ \\
    \midrule
    TC1 (full) & \hspace{-2mm}28,642 & 0.5271 & 0.5271 \\
    \midrule
    TC2 (linear) & 8 & 0.5287 & 0.5313 \\
    TC2 (linear) & 32 & 0.5279 & 0.6477 \\
    TC2 (linear) & 128 & 0.5273 & 0.5274 \\
    TC2 (linear) & 512 & 0.5272 & 0.5272 \\
    \midrule
    TC3 (tanh map) & 8 & 1.4401 & 1.4617 \\
    TC3 (tanh map) & 32 & 0.7668 & 0.9254 \\
    TC3 (tanh map) & 128 & 0.5406 & 0.5539 \\
    TC3 (tanh map) & 512 & 0.5279 & 0.5283 \\
    \midrule
    TC4 (last layer) & 578 & 0.5271 & 0.5271 \\
    \bottomrule
    \end{tabular}
\end{wraptable}

\vspace{1mm}\noindent\textbf{The nonlinear map.} \looseness=-1 TC3 is dominated by TC2 at every matched $d$ and is far slower to converge at $d = 8$ (median $\varepsilon = 1.46$ within 880 steps, Fig.~\ref{fig:fourarm}). The nonlinear construction therefore underperforms on the frustrated system, whose sign-free optimum is a set of classical configurations. On the TFIM it is roughly on par with the linear subspace (Appendix~\ref{app:tc3tfim}). We tested one nonlinear construction, a tanh variant of the mapping of \citet{sen2026mapping} trained with the task loss only, without their weight modulation or auxiliary losses, and in one architecture family. Our statements about the nonlinear map are limited to this variant.

\vspace{1mm}\noindent\textbf{Seed dependence and conditioning.} \looseness=-1 The seeds of TC2 are bimodal. They end at $\varepsilon \approx 0.527$ or at $0.646$, that is, at $E \approx -4$ or $-3$, which are both energies of classical Ising configurations. The \emph{orientation} of the random subspace selects the basin, which is why $d^*$ is defined by a median over seeds. At $d = 512$ all six runs reach the lower attractor (Appendix~\ref{app:tc3tfim}). The condition number of the QGT in $z$-space does not predict the quality of the optimization. TC2 wins with a condition number of about $10^{10}$, and TC3 loses with $10^{2}$ to $10^{4}$. The conditioning of the curvature is therefore a poor proxy for quality here.

\subsection{An exact-sign control separates amplitude from sign}
\label{sec:sign}

\looseness=-1 This subsection asks where the difficulty of the ground state lies, in the amplitude or in the sign. Section~\ref{sec:fourarm} measured $d^*$ against the positive-cone bound. To get below the bound the model must represent signs, so we extend the same ansatz with a small \emph{phase network}, $\log\psi = \log|\psi| + i\phi(\sigma)$. Here $\phi$ is a 2-layer tanh MLP. It holds 1,633 of the $P_c = 30{,}275$ real parameters of the extended model (5.4\%). We call these parameters the sign block and the rest the amplitude block. All runs use the same coupling ($4\times4$, $J_2/J_1 = 0.5$) and 3 seeds per cell, and they log 0 divergence events.

\vspace{1mm}\noindent\textbf{Full-parameter reference.} \looseness=-1 Configurations on the sign-capable model carry the suffix \emph{c}. Full-parameter training of the extended model, TC1c, reaches a median best $\varepsilon$ of 0.0444 (seeds 0.0382, 0.0444 and 0.1920), which is $12\times$ below the bound. It does so in only ${\sim}600$ steps.

\vspace{1mm}\noindent\textbf{Proportional allocation fails.} \looseness=-1 The random-subspace construction that reaches the bound of Sec.~\ref{sec:fourarm} at $d = 8$ stays far from the sign-capable reference. The medians of TC2c are 0.690, 0.482, 0.395 and 0.316 at $d = 8$, 32, 128 and 512, against the pre-registered gate of $2\times$TC1c $= 0.0888$, even though these runs take 1,420--2,040 steps against the 600 of the reference. This failure has two possible causes. Proportional allocation assigns latent dimensions by group size, so the phase network receives only ${\sim}28$ of 512 dimensions, and the sign block may simply have too few dimensions. Alternatively, the sign structure itself may need many dimensions. Reallocating within one budget cannot separate the two, because every dimension moved into the sign block leaves the amplitude block. We therefore hold one block fixed and sweep the other, and add an exact-sign control on the amplitude side.

\begin{figure}[t]
\centering
\includegraphics[width=0.92\linewidth]{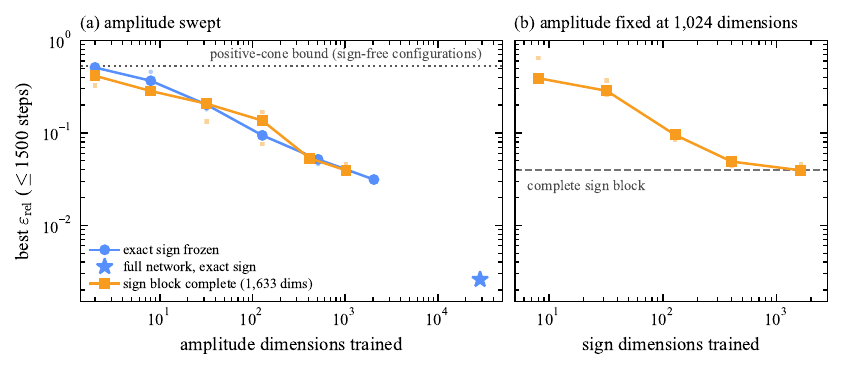}
\caption{\textbf{Sign/amplitude dissection in one sign-capable model} ($4\times4$, $J_2/J_1 = 0.5$; best $\erel$ within 1,500 steps; lines are medians of 3 seeds, small dots are seeds). \captiona\ Amplitude swept with the sign block complete (orange), beside the exact-sign control (blue; the star is the full network, which completed 1,380--1,460 steps) and the positive-cone bound, below which no sign-free configuration of Table~\ref{tab:fourarm} can go (dotted). \captionb\ Sign swept with the amplitude fixed at 1,024 dimensions; dashed: the complete sign block.} %
\label{fig:split}
\end{figure}

\vspace{1mm}\noindent\textbf{The exact-sign control.} \looseness=-1 We freeze the exact sign of the ground state, taken from exact diagonalization, into the amplitude network, $\log\psi = \log|\psi_\theta| + i\pi\,[s_0(\sigma) < 0]$. Every trainable direction is then an amplitude direction, and the target is the true ground state. The sign is a lookup table and adds no parameters. Each run takes 1,500 steps, with 3 seeds and the same sampler and optimizer. The amplitude is not cheap. The median $\erel$ falls smoothly from 0.37 at $d = 8$ to 0.052 at $d = 512$ and 0.031 at $d = 2048$, while the full amplitude network reaches $2.6\times10^{-3}$ (Fig.~\ref{fig:split}a). The $d = 8$ cells have converged, with the best $\erel$ within 3.1\% over their last 300 steps, so the $3{,}500\times$ of Sec.~\ref{sec:fourarm} does not survive once the bound is removed. No subspace up to $d = 2048$ reaches $10^{-2}$.

\vspace{1mm}\noindent\textbf{Amplitude swept, sign block complete.} \looseness=-1 With all 1,633 dimensions of the sign block trainable, the error falls from 0.42 at 2 amplitude dimensions to 0.053 at 415 and 0.039 at 1,024 (Fig.~\ref{fig:split}a, and Table~\ref{tab:alloc} in the appendix). From 415 dimensions on it matches the exact-sign control within 6\%. At 2 and 8 it does better than the exact sign, which need not be the best sign for a restricted amplitude.

\vspace{1mm}\noindent\textbf{Sign swept, amplitude fixed.} \looseness=-1 With the amplitude fixed at 1,024 dimensions, the median error at 402 of the 1,633 sign dimensions is $1.24\times$ that of the complete block. At 128 it is $2.4\times$ and at 32 it is $7.3\times$ (Fig.~\ref{fig:split}b). At this amplitude dimension the seed ranges at 402 and 1,633 sign dimensions overlap (0.044--0.050 against 0.038--0.046). At 415 amplitude dimensions they separate (0.060--0.067 against 0.050--0.055, a ratio of $1.18\times$). Once the amplitude is resolved, the sign therefore needs its whole block. The full grid (Fig.~\ref{fig:heatmap} in the appendix) shows that this demand appears only then. With 2--128 amplitude dimensions, a 402-dimensional sign block comes within $1.1\times$ of the complete one ($0.73$--$1.07\times$). Proportional allocation leaves the sign block 28 dimensions, and it sits $5.9\times$ above the exact-sign control at the same 484 amplitude dimensions, both within 1,420 steps.

\vspace{1mm}\noindent\textbf{The cost of the sign in optimization.} \looseness=-1 With every parameter trainable, the sign-capable model reaches a median $\erel$ of 0.044 in its first 600 steps, against $4.9\times10^{-3}$ for the same amplitude network given the exact sign. This is a factor of 9 at equal steps under the same minSR solver. It fits an analysis of estimator variance in which the phase gradient can bottleneck the training of complex-valued NQS \citep{xue2026lowvariance}.

\vspace{1mm}\noindent\textbf{Dependence on frustration.} \looseness=-1 We scan the frustration $J_2/J_1 \in \{0.40, 0.45, \ldots, 0.70\}$ at fixed $d$, at both lattice sizes (Table~\ref{tab:signlambda} and Fig.~\ref{fig:signlambda} in Appendix~\ref{app:figures}). The error follows the phase structure of the model. At $4\times4$, at $d = 128$ and $d = 512$, every coupling with $J_2/J_1 \le 0.55$ has a higher error than every coupling with $J_2/J_1 \ge 0.60$, and at $6\times6$ the error at $d = 3072$ peaks at $J_2/J_1 = 0.55$.

\vspace{1mm}\noindent\textbf{The saturated allocation at $6\times6$.} \looseness=-1 At $6\times6$ the allocation $d = 3072$ gives the sign block all of its 2,273 dimensions and the amplitude the remaining 799. Its median per coupling still stays $3.8$--$7.7\times$ above the best full-parameter seed. We gate on the best seed here and ask the median to come within $2\times$ of it, which we call the robust gate (Sec.~\ref{sec:breakage}). Under this gate $d^*_c$, the smallest $d$ that passes the gate, is right-censored ($> 3072$) at every coupling. The sign block is fully covered while the amplitude has only 799 of 28,642 dimensions, so the two deficits cannot be separated here. We report a joint failure of the saturated budget (Appendix~\ref{app:gates}).

\subsection{$d^*$ tracks the phase transition, above a floor set by the random slice}
\label{sec:tfim}

\looseness=-1 This subsection asks what the measured dimension responds to. We measure $d^*$ across a quantum phase transition, where the structure of the ground state changes, and in a limit where the ground state is nearly trivial.

\noindent\textbf{Setup.} \looseness=-1 We use TFIM chains. Their ground states have no signs, so the real-amplitude ARNN is adequate and small thresholds are meaningful. The linear subspace (TC2) is the primary configuration. We use three protocols and never mix them. The base protocol uses $L = 16$, a fixed budget of 2000 steps and $d \in \{2, \ldots, 128\}$. The deep protocol uses $L = 16$ and 6000 steps, a $3\times$ larger budget, with $d$ extended to 256 in the critical region (0 unhealthy cells). The third is an $L = 24$ grid under the deep protocol with $d$ up to 512 (0 unhealthy). The field grids straddle the exact critical point $h_c/J = 1$. The deep protocol continues into the paramagnet at $h/J \in \{4, 10\}$, with $d \in \{32, 64, 128\}$ (0 divergence events) and a full-parameter reference with 3 seeds per field.

\begin{wrapfigure}{r}{0.48\textwidth}
\centering
\includegraphics[width=1\linewidth]{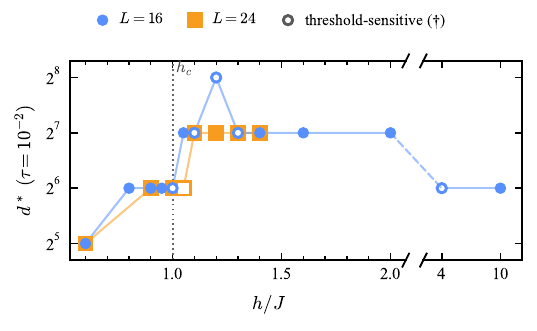}
\caption{$d^*(h)$ at $\tau = 10^{-2}$ for TFIM chains at $L = 16$ and $L = 24$ (deep protocol, Table~\ref{tab:staircase}). The demand for trainable dimensions rises ${\sim}4\times$ across the quantum phase transition at both sizes. Deep in the paramagnet ($L = 16$, broken axis) it relaxes only to 64 dimensions, although at $h/J = 10$ a product state without parameters is within $\erel = 2.5\times10^{-3}$ of the ground-state energy.}
\label{fig:staircase}
\end{wrapfigure}

\noindent\textbf{The rise.} \looseness=-1 At both sizes $d^*$ rises ${\sim}4\times$ across the transition (Fig.~\ref{fig:staircase}). The edge of the step lies inside the critical window, between $h/J = 1.0$ and $1.05$ for $L = 16$ and between $1.05$ and $1.1$ for $L = 24$. Table~\ref{tab:staircase} in Appendix~\ref{app:figures} lists the value in every cell. The demand stays elevated through $h/J = 2.0$ and relaxes deep in the paramagnet, to $d^* = 64$ at both $h/J = 4$ and $10$.

\noindent\textbf{The floor.} \looseness=-1 The relaxation stops short of what the state needs. At $h/J = 10$ the product state $|{\rightarrow}\rangle^{\otimes L}$, which has no parameters, is already within $\erel = 2.5\times10^{-3}$ of the exact energy, and the full-parameter reference reaches the Monte Carlo noise floor. A random 32-dimensional slice of the same network stalls at a median of $3.4\times10^{-2}$, which is 14 times worse than the parameter-free state. The stall is converged, since the best $\erel$ is unchanged over the final 20\% of steps in all six $d = 32$ cells at $h/J \in \{4, 10\}$. The plateau shows the same split. The full-parameter reference is at its best at $h/J = 2.0$, with a median $\erel$ of $2.5\times10^{-3}$, and its medians span $[2.5, 4.1]\times10^{-3}$ over $h/J \in \{1.0, 1.6, 2.0\}$ with 3 seeds each and zero divergence events. At the same field, $d = 64$ subspaces still fail, with a median of $1.16\times10^{-2}$. $d^*$ therefore tracks the transition above a floor, and the floor belongs to the random slice rather than to the state.

\noindent\textbf{Independence of the budget.} \looseness=-1 Under a fixed step budget, a larger $d^*$ could mean that more directions are required, or that the same directions take longer to find. The pair of protocols separates the two. At $\tau = 10^{-2}$, all 7 cells that the base and deep protocols share are unchanged under the $3\times$ larger step budget, and 0 cells shrank, so more steps do not lower $d^*$ (Fig.~\ref{fig:protocol} in the appendix). The same control \emph{fails} at tighter thresholds, which is why every $d^*$ claim in this paper is pinned to $\tau = 10^{-2}$.

\noindent\textbf{Pre-registered falsification.} \looseness=-1 The $L = 16$ staircase contains an interior maximum, $d^*(1.2) = 256$, whose deciding median sits within $+4\%$ of the threshold. We pre-registered competing hypotheses for the $L = 24$ grid, and the hypothesis that it is an artifact was selected. At $L = 24$ \textbf{no coupling requires $d = 256$}, so the step is what survives (Appendix~\ref{app:figures}).

\noindent\textbf{Dependence on the threshold.} \looseness=-1 We sweep $\tau$ over $[5\times10^{-3}, 2\times10^{-2}]$ on the existing runs. The \emph{location} of the step drifts with the threshold, while its \emph{existence and confinement} do not. Wherever a step exists, its edge lies inside the finite-size critical window $h/J \in [0.9, 1.2]$ at both sizes (Fig.~\ref{fig:fan} in the appendix).

\subsection{Full-parameter training breaks at the larger lattice}
\label{sec:breakage}

\looseness=-1 \textbf{The full-parameter reference broke, and the subspace configurations did not.}

\looseness=-1 We train the sign-capable model with every parameter trainable, under the same SGD and minSR configuration as everywhere else, with 5 seeds per coupling. At $6\times6$ its final energy is bimodal across seeds (Fig.~\ref{fig:stab66} in the appendix). At 4 of 7 couplings the median seed ends in a catastrophic attractor, which we call the bad basin, with $E \approx -11$ to $-14$ against exact energies of $-17.8$ to $-19.1$, computed by Lanczos diagonalization in a sector of 126 million states (Appendix~\ref{app:ed66}). The 35 runs of this reference logged 41 divergence events and 3 aborts.

\looseness=-1 All 70 subspace runs at this size completed with zero divergence events. Their seed spread is $1.0$--$3.3\times$ per two-seed cell, against $10$--$52\times$ across the five seeds of TC1. For a comparison at equal steps we cut every subspace run at the median step count of TC1, which is 1,000 at every coupling. The $d = 3072$ subspace then \emph{beats the full-parameter median at exactly those four couplings}, with per-coupling ratios of the median $\erel$ of $1.6$--$4.6\times$ in its favour. It loses by $2.3$--$5.0\times$ at the three couplings where the reference median stays out of the bad basin ($J_2/J_1 = 0.45$--$0.55$). The low-dimensional slice through $\theta_0$ removes the divergence but not the bad basin. When we rerun the same subspace with the minSR preconditioner of TC1, it is itself bimodal across seeds and still logs no divergence event (Sec.~\ref{sec:geometry}).

\looseness=-1 With the median seed in the bad basin, a gate that refers to the median of the reference loses its meaning, which is why Sec.~\ref{sec:sign} gates against $2\times$ the \emph{best} reference seed. Appendix~\ref{app:gates} gives both gates. We observed this failure for one optimizer configuration, plain SGD under SR or minSR. We do not claim that full-parameter NQS training is unstable in general, only that the subspace configuration is the stable one at matched optimizer and budget.

\section{Discussion}
\label{sec:discussion}

\subsection{Proportional allocation and small parameter blocks}
\label{sec:peft}

\looseness=-1 PEFT methods such as LoRA give every adapted weight matrix the same rank, so a block's budget follows its shape and not its demand \citep{hu2022lora}; AdaLoRA reallocates it by importance \citep{zhang2023adalora}. Our dissection is a \emph{measured} instance of how this can fail. The sign block holds 5.4\% of the parameters, so proportional allocation gives it 28 of 512 dimensions, which leaves the error $5.9\times$ above an exact-sign control at the same amplitude allocation. Yet once the amplitude is well resolved, even a quarter of the sign block leaves the median error $1.24\times$ higher (Table~\ref{tab:alloc}). Low-rank adaptation with budgets set by shape gives too little to any small block whose demand exceeds its share.

\subsection{Implications for NQS practice}
\label{sec:practice}

\looseness=-1 First, \textbf{subspace training is a stabilizer}. We saw zero divergences across the 246 subspace runs of the sign-structure study and in every deep-protocol and $L=24$ TFIM cell, while full-parameter training becomes bimodal at $6\times6$. It is worth considering wherever full-parameter VMC is fragile, within the scope stated in Appendix~\ref{app:gates}. Second, \textbf{$d^*$ is a cheap probe with a floor}. Like the compute exponent of NQS scaling laws \citep{rende2026scaling}, it responds to the difficulty of the coupling, at roughly two orders of magnitude less compute per coupling, but its value includes what the random slice itself costs. Third, the \textbf{variational-floor guard} (Sec.~\ref{sec:health}) should be standard in any VMC that trains a latent reparameterization. It costs nothing whenever $E_0$ or a good lower bound is known.

\section{Conclusion}
\label{sec:conclusion}

\looseness=-1 We brought the measurement of intrinsic dimension to variational Monte Carlo. A neural quantum state is trained inside a random subspace of its weights, SR applies without modification, and every run is scored against an exact energy. Three findings follow. Subspace training is robust under gradient noise. In our sign-structure study it never diverged, while full-parameter training with the same optimizer settings diverged repeatedly at the larger lattice. Apparent compression can be a variational bound. A network with non-negative amplitudes reaches its best energy in 8 random directions, but no such network can reach a lower energy. Once the model can represent signs, neither the sign nor the amplitude is cheap. The measured dimension tracks a phase transition, above a floor set by the random slice, so part of $d^*$ is a cost of the slice itself. Two measurements are the natural next steps. The first is to repeat the exact-sign control in other network families, since all our results use one. The second is to compare $d^*$ with compute scaling laws \citep{rende2026scaling} on a shared Hamiltonian.

\subsection*{AI use statement}

We used a large language model to polish the writing. We have reviewed all AI-assisted work and take full responsibility for the content of this paper.

\bibliography{references}
\bibliographystyle{iclr2027_conference}

\appendix
\section{Both gates and the scope of the $6\times6$ failure}
\label{app:gates}

This appendix gives both gates, the prescription that we draw from them, and the scope of the failure of full-parameter training.

\vspace{1mm}\noindent\textbf{Median-referenced gates do not survive scale-up.} A gate is the error that a subspace run must reach, and we set it as a multiple of the error of the full-parameter reference. With the median seed in the bad basin, a gate at $2\times$ the median of the reference inflates until it carries no information (0.31--0.57 on the final 5-seed data). The $d^*_c$ values that it produces are artifacts of this inflation, and we discard them. We report the robust gate instead, which is $2\times$ the \emph{best} reference seed. Under it $d^*_c$ at $6\times6$ is right-censored ($> 3072$) at every coupling (Sec.~\ref{sec:sign}). We state both gates, although reporting the median gate alone would have been more favorable to the subspace. The prescription is general. When the reference configuration can be multi-modal across seeds, gate against a robust statistic of the reference, or repair the reference before measuring against it.

\vspace{1mm}\noindent\textbf{The saturated allocation at $6\times6$.} At $6\times6$ the sign block holds 2,273 of $P_c = 30{,}915$ parameters. The saturated allocation, $d = 3072$, gives the sign block all of its 2,273 dimensions and the amplitude the remaining 799. Its median per coupling still stays $3.8$--$7.7\times$ above the best full-parameter seed, with a per-seed range of $2.4$--$8.1\times$. We gate on the best seed here and ask the median to come within $2\times$ of it, which we call the robust gate (Sec.~\ref{sec:breakage}). Under this gate the demand of the sign-capable model, $d^*_c$, defined as the smallest $d$ that passes the gate, is right-censored ($> 3072$) at every coupling. This locates a failure but does not attribute it to the sign. At $d = 3072$ the sign block is \emph{fully} covered, so the residual error cannot show that the signs need more than their own 2,273 directions. At the same time the amplitude block has only 799 of 28,642 dimensions, and the two deficits cannot be separated at this allocation. We therefore report the $6\times6$ result as a joint failure of the saturated budget.

\begin{figure}[H]
\centering
\includegraphics[width=0.62\linewidth]{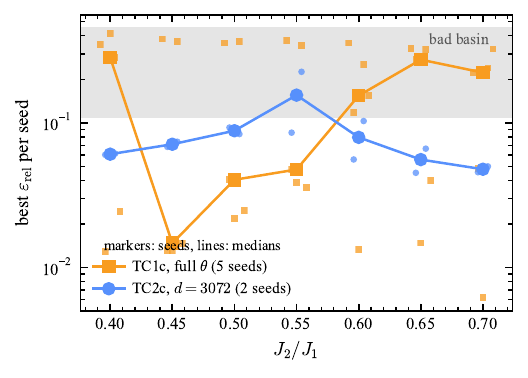}
\caption{Every seed of the $6\times6$ sign scan. Best relative error per seed against $J_2/J_1$ for the full-parameter reference TC1c (5 seeds per coupling) and the saturated subspace TC2c at $d = 3072$ (2 seeds), scored against the exact energies of Appendix~\ref{app:ed66}. At four couplings the median TC1c seed sits in the bad basin (shaded), and its seeds spread over more than an order of magnitude; the two subspace seeds stay within $3.3\times$ of each other. The subspace runs logged no divergence event, the reference 41 events and 3 aborts (Sec.~\ref{sec:breakage}).}
\label{fig:stab66}
\end{figure}

\vspace{1mm}\noindent\textbf{Scope.} We demonstrated this failure of full-parameter training for one optimizer configuration. It is plain SGD under SR or minSR, the uniform choice that makes our configurations comparable. We do not claim that full-parameter NQS training is unstable in general, and better-tuned or adaptive full-parameter optimizers may well close the gap. The controlled comparison shows that \emph{at matched optimizer and budget} the subspace configuration is the stable one.

\section{The variational-floor incident and audit}
\label{app:floor}

This appendix gives the four criteria of the health protocol, the incident that motivated the variational-floor guard, and the audit that followed.

An event is declared when the Monte Carlo energy estimate $E$ meets any of four conditions. (1) $E$ is not finite. (2) $E$ jumps upward, $E > E_{\mathrm{best}} + 2.0$, where $E_{\mathrm{best}}$ is the best energy so far. (3) $E$ violates the \textbf{variational floor}, $E < E_0 - \max(0.5,\ 5\sigma_E)$, when $E_0$ is known. (4) $E$ jumps downward, $E < E_{\mathrm{best}} - 20$, when $E_0$ is not known. Guards (3) and (4) are necessary when the objective is noisy. Milder violations of the floor would go unnoticed and look like convergence.

One run in our TFIM grid recorded a ``best'' energy of $-4.7\times10^{7}$. This is far below the variational floor $E_0$, which no legitimate variational state can cross. Four cascading rollbacks followed, each \emph{to the poisoned state}, and then an abort. Guards (1) and (2), which detect NaNs and upward jumps, saw nothing wrong, because the corruption moved the estimate \emph{downward}. A milder violation of the floor would have passed for convergence without any warning. After the fix we audited all 147 base-protocol runs. Exactly one run was below the floor, the known one, so there was no silent contamination. We re-ran that seed.

\section{Exact reference energies at $6\times6$}
\label{app:ed66}

The $6\times6$ reference energies are exact. We diagonalize the spin-$\tfrac12$ $J_1$--$J_2$ Hamiltonian on the $6\times6$ torus in the $S^z = 0$ sector. We reduce this sector by the two lattice translations at momentum zero and by spin inversion with eigenvalue $+1$, which gives the sector of the singlet ground state. No point-group symmetry is used, so every rotation sector is included and the level crossing of the collinear regime cannot be missed. The sector holds 126,056,625 states, and the Hamiltonian has $9.3\times10^{9}$ nonzero entries. With the full point group the same lattice reduces to 15.8 million states \citep{schulz1996magnetic}, and \citet{richter2010spin} reach $N = 40$ in a sector of 430,909,650 states. A plain Lanczos recursion converges the lowest eigenvalue to $10^{-12}$ relative in 58--118 steps. This takes about ten minutes per coupling on one CPU node, and the basis and the matrix structure are shared by all couplings.

We made three checks. At $4\times4$ the same code reproduces the full-space diagonalization of the main text at all seven couplings to $3\times10^{-13}$. At $4\times8$ it agrees with an independent implementation in QuSpin \citep{weinberg2019quspin} to $3\times10^{-14}$. At $6\times6$ the energies per site at $J_2/J_1 = 0.5$ and $0.55$, $-0.503810$ and $-0.495178$, are the $N = 36$ values of \citet{schulz1996magnetic}.

Table~\ref{tab:ed66} lists all seven energies beside DMRG torus references computed with TeNPy \citep{hauschild2024tenpy} ($\chi = 4096$). The DMRG energies are variational upper bounds. They lie above the exact energy by $1.6\times10^{-5}$ to $1.5\times10^{-3}$ relative, and the gap is largest in the frustrated window.

\begin{table}[H]
\caption{Exact ground-state energies of the $6\times6$ $J_1$--$J_2$ torus beside the DMRG references ($\chi = 4096$).}
\label{tab:ed66}
\begin{center}
\small
\begin{tabular}{lcccc}
\toprule
$J_2/J_1$ & $E_0$ (exact) & $E_0/N$ & $E_{\mathrm{DMRG}}$ & $(E_{\mathrm{DMRG}} - E_0)/|E_0|$ ($10^{-4}$) \\
\midrule
0.40 & $-19.0708202$ & $-0.5297450$ & $-19.0604506$ & 5.4 \\
0.45 & $-18.5636661$ & $-0.5156574$ & $-18.5496834$ & 7.5 \\
0.50 & $-18.1371475$ & $-0.5038097$ & $-18.1179292$ & 10.6 \\
0.55 & $-17.8263972$ & $-0.4951777$ & $-17.8000652$ & 14.8 \\
0.60 & $-17.7565894$ & $-0.4932386$ & $-17.7403613$ & 9.2 \\
0.65 & $-18.2371629$ & $-0.5065879$ & $-18.2368670$ & 0.16 \\
0.70 & $-19.0800441$ & $-0.5300012$ & $-19.0787294$ & 0.69 \\
\bottomrule
\end{tabular}
\end{center}
\end{table}

\section{More on the four training configurations}
\label{app:tc3tfim}

\noindent\textbf{Reruns of the $4\times4$ seeds.} Which attractor a seed reaches varies between reruns. At equal steps, two of the three $d = 8$ seeds in Table~\ref{tab:fourarm} reach the lower attractor, against one of three at $d = 32$. In a fresh 1000-step rerun of the same seeds, none of them reaches it at either $d$ (medians 0.717 and 0.646).

\noindent\textbf{The nonlinear map on the TFIM.} The base protocol also ran TC3 alongside TC2 at matched $d$. On the TFIM the tanh mapping is roughly comparable to the linear subspace, and it is better at $d = 2$ and $h/J = 0.6$, where the best $\varepsilon$ of TC3 is $5.6\times10^{-2}$ against a TC2 median of $2.6\times10^{-1}$. Together with Sec.~\ref{sec:fourarm}, this shows that the weakness of this mapping variant is specific to the frustrated system and is not caused by VMC noise as such.

\begin{figure}[H]
\centering
\includegraphics[width=0.6\linewidth]{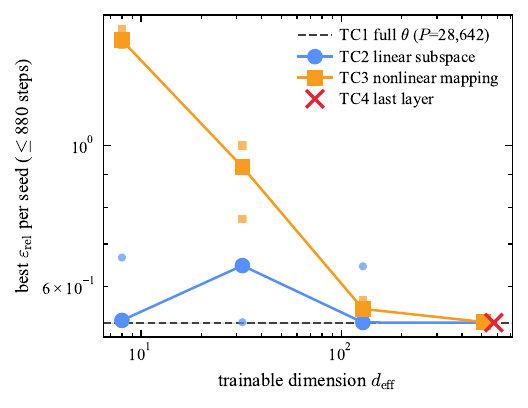}
\caption{The plot shows best $\erel$ within the first 880 steps against trainable dimension $d$ for the four configurations of Table~\ref{tab:fourarm}, at $4\times4$ and $J_2/J_1 = 0.5$. Small dots are single seeds, and the lines are medians over 3 seeds. The linear subspace (TC2) reaches the positive-cone bound shared by every sign-free configuration already at $d = 8$. The nonlinear map (TC3) lies above the linear curve at every matched $d$ and is slowest at $d = 8$.}
\label{fig:fourarm}
\end{figure}

\section{Additional figures and tables}
\label{app:figures}

\begin{figure}[H]
\centering
\includegraphics[width=0.8\linewidth]{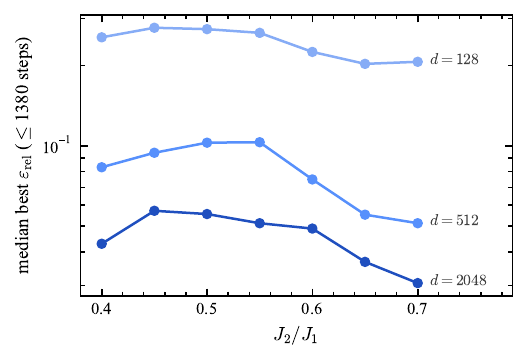}
\caption{Error of the sign-capable model against frustration $J_2/J_1$ at fixed subspace dimension, on the $4\times4$ $J_1$--$J_2$ lattice (Sec.~\ref{sec:sign}, Table~\ref{tab:signlambda}). The plot shows the median best $\erel$ within 1,380 steps over 2 seeds, with phase-weighted allocation at $d \in \{128, 512, 2048\}$. At $d = 128$ and $d = 512$ every coupling with $J_2/J_1 \le 0.55$ has a higher error than every coupling with $J_2/J_1 \ge 0.60$. At $d = 2048$ the two ranges overlap.}
\label{fig:signlambda}
\end{figure}

\noindent\textbf{Dependence on frustration.} We scan the frustration $J_2/J_1 \in \{0.40, 0.45, \ldots, 0.70\}$ with phase-weighted subspaces at $d \in \{128, 512, 2048\}$ and a full-parameter reference, with 2 seeds each (56 runs, 0 failures). Phase-weighted allocation weights the share of the sign block in the latent budget by 64 times its parameter count, so the sign block fills first and holds 1,606 of its 1,633 dimensions at $d = 2048$. The error at fixed $d$ follows the phase structure of the model (Table~\ref{tab:signlambda} and Fig.~\ref{fig:signlambda}). With every subspace run cut at 1,380 steps, the $d = 512$ median is 0.083--0.103 for $J_2/J_1 \le 0.55$ and 0.051--0.075 for $J_2/J_1 \ge 0.60$. The $d = 128$ median splits the same way (0.256--0.278 against 0.203--0.225), while at $d = 2048$ the two ranges overlap (0.043--0.057 against 0.031--0.049). We report no gated dimension at $4\times4$. The two-seed full-parameter reference is bimodal across seeds here, as it is at $6\times6$, and rerunning its seeds moves them between basins. At $J_2/J_1 = 0.40$ its median moves $3\times$ at equal steps, so a gate built on it is not stable. At $6\times6$ (105 runs, scored against exact energies, Appendix~\ref{app:ed66}) the same picture holds at fixed $d$. The error at $d = 3072$ peaks at $J_2/J_1 = 0.55$, in the N\'eel--frustrated region, and relaxes $3\times$ into the striped side, which matches the $4\times4$ curves qualitatively.

\begin{table}[H]
\caption{Error of the sign-capable model vs.\ frustration at fixed subspace dimension. $4\times4$ rows: median best $\varepsilon$ within 1,380 steps (phase-weighted allocation, 2 seeds). $6\times6$ row: median $\varepsilon$ at fixed $d = 3072$, the saturated allocation at that size.}
\label{tab:signlambda}
\begin{center}
\footnotesize
\begin{tabular}{lccccccc}
\toprule
$J_2/J_1$ & 0.40 & 0.45 & 0.50 & 0.55 & 0.60 & 0.65 & 0.70 \\
\midrule
$4\times4$, $d = 128$ & 0.256 & 0.278 & 0.274 & 0.266 & 0.225 & 0.203 & 0.207 \\
$4\times4$, $d = 512$ & 0.083 & 0.094 & 0.103 & 0.103 & 0.075 & 0.055 & 0.051 \\
$4\times4$, $d = 2048$ & 0.043 & 0.057 & 0.056 & 0.051 & 0.049 & 0.037 & 0.031 \\
$6\times6$, $d = 3072$ & 0.061 & 0.071 & 0.088 & \textbf{0.156} & 0.079 & 0.056 & 0.048 \\
\bottomrule
\end{tabular}
\end{center}
\end{table}

\begin{table}[H]
\caption{Sign/amplitude dissection in one sign-capable model ($4\times4$, $J_2/J_1 = 0.5$; median best $\erel$ within 1,500 steps, 3 seeds). Top: sign block complete, amplitude swept, beside the exact-sign control at the same amplitude dimension (interpolated log-log at 415 and 1,024). Bottom: amplitude fixed at 1,024 dimensions, sign swept, as a ratio to the complete block. \emph{From 415 amplitude dimensions on the learned sign matches the exact one. With the amplitude this well resolved, a quarter of the sign block already leaves the median error $1.24\times$ higher.}}
\label{tab:alloc}
\begin{center}
\footnotesize
\begin{tabular}{rrcr}
\toprule
Sign  & Ampl.  & median  & Exact sign  \\
 dims &  dims &  $\varepsilon$ &  (ratio) \\
\midrule
1,633 & 2 & 0.415 & 0.511 ($0.81\times$) \\
1,633 & 8 & 0.284 & 0.367 ($0.77\times$) \\
1,633 & 32 & 0.207 & 0.202 ($1.02\times$) \\
1,633 & 128 & 0.136 & 0.094 ($1.44\times$) \\
1,633 & 415 & 0.053 & 0.057 ($0.94\times$) \\
1,633 & 1,024 & 0.039 & 0.040 ($0.98\times$) \\
\midrule
& & & Ratio to  \\
& & &  complete block \\
402 & 1,024 & 0.049 & $1.24\times$ \\
128 & 1,024 & 0.095 & $2.42\times$ \\
32 & 1,024 & 0.286 & $7.25\times$ \\
8 & 1,024 & 0.391 & $9.93\times$ \\
\bottomrule
\end{tabular}
\end{center}
\end{table}

\begin{figure}[H]
\centering
\includegraphics[width=0.8\linewidth]{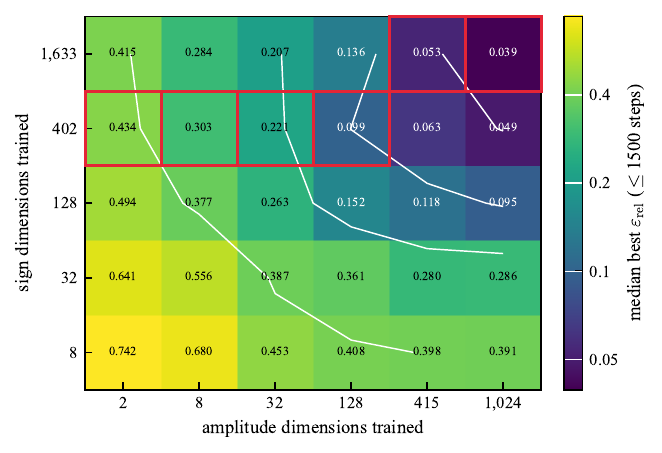}
\caption{The sign/amplitude grid behind Table~\ref{tab:alloc} and Fig.~\ref{fig:split}, for one sign-capable model at $4\times4$ and $J_2/J_1 = 0.5$. Each cell shows the median best $\erel$ within 1,500 steps over 3 seeds, and the white contours lie at $\erel = 0.05$, $0.1$, $0.2$ and $0.4$. In each amplitude column a red outline marks the smallest sign block whose median is within $1.1\times$ of that of the complete block. With 2--128 amplitude dimensions a 402-dimensional sign block is enough. From 415 amplitude dimensions on, the sign needs its whole block.}
\label{fig:heatmap}
\end{figure}

\begin{figure}[H]
\centering
\includegraphics[width=0.8\linewidth]{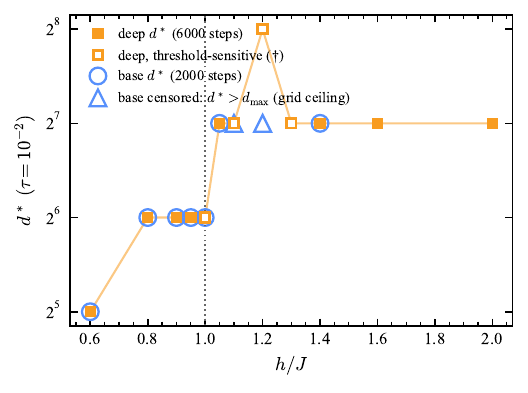}
\caption{Control for the independence of the $d^*(h)$ staircase from the step budget (Sec.~\ref{sec:tfim}). The base protocol (TFIM, $L = 16$, fixed 2000-step budget) is compared with the deep protocol (6000 steps, $3\times$, with $d$ extended to 256 in the critical region). At $\tau = 10^{-2}$, all 7 cells that the two protocols share are unchanged under the $3\times$ budget, and 0 cells shrank. The two cells that are censored under the base protocol resolve within the deep grid ($h/J = 1.1 \to 128$, $1.2 \to 256$). At this threshold the staircase is therefore not limited by the budget. At tighter thresholds the control fails. At $3\times10^{-3}$ the outcome is mixed, and at $10^{-3}$ the one comparable cell moved $128 \to 64$. This is why we restrict claims about $d^*$ to $\tau = 10^{-2}$.}
\label{fig:protocol}
\end{figure}

\noindent\textbf{Independence of the budget.} Under a fixed step budget, a larger $d^*$ could mean that more directions are required, or that the same directions take longer to find. The pair of protocols separates the two. At $\tau = 10^{-2}$, all 7 cells that the base and deep protocols share are unchanged under the $3\times$ larger step budget, and 0 cells shrank. The two cells that were censored in the base protocol resolved within the deep grid ($h/J = 1.1 \to 128$ and $1.2 \to 256$). More steps therefore do not lower $d^*$, and the cells that lacked an answer needed more directions (Fig.~\ref{fig:protocol}). The same control \emph{fails} at tighter thresholds. At $3\times10^{-3}$ the outcome is mixed, and at $10^{-3}$ the one comparable cell moved $128 \to 64$. This is why every $d^*$ claim in this paper is pinned to $\tau = 10^{-2}$. The threshold at which $d^*$ is independent of the budget is thus measured and not assumed.

\noindent\textbf{Pre-registered falsification.} The $L = 16$ staircase contains an interior maximum, $d^*(1.2) = 256$. We resolved it with $n = 5$ seeds, which give a median $\varepsilon$ of 0.01039 with 4/5 seeds above the threshold. The maximum is therefore confirmed at this size, but we flag it with $\dagger$ because the deciding median sits within $+4\%$ of the threshold. Rather than interpret it as physics, we pre-registered competing hypotheses for the $L = 24$ grid. The hypothesis that it is an artifact was selected. At $L = 24$ \textbf{no coupling requires $d = 256$}, and the cell that sat 4\% above the threshold at $L = 16$ reads $8.0\times10^{-3}$ at $L = 24$. The interior maximum is thus an artifact of the finite size and of the discrete threshold, and the step is what survives.

\begin{figure}[H]
\centering
\includegraphics[width=0.7\linewidth]{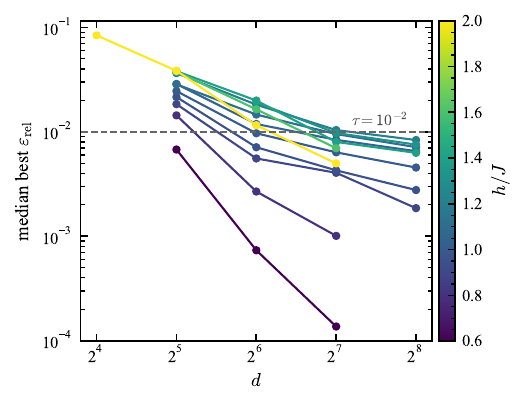}
\caption{Median $\erel$ against trainable dimension $d$, with one curve per transverse field $h$ (TFIM, $L = 16$, deep protocol, Sec.~\ref{sec:tfim}). The $d^*(h)$ staircase of Fig.~\ref{fig:staircase} is a level cut of this smooth error surface $\varepsilon_{\mathrm{med}}(h, d)$ at the threshold $\tau = 10^{-2}$. Across the scanned window $h/J \le 2$, larger fields beyond the transition need a systematically larger $d$ to reach the same accuracy.}
\label{fig:fan}
\end{figure}

\noindent\textbf{Dependence on the threshold.} We sweep $\tau$ over $[5\times10^{-3}, 2\times10^{-2}]$ on the existing runs, which needs no new compute. The \emph{location} of the step drifts with the threshold, while its \emph{existence and confinement} do not. Wherever a step exists, its edge lies inside the finite-size critical window $h/J \in [0.9, 1.2]$ at both sizes. The ordered side below the window ($h/J \le 0.8$) is stable in threshold and size, with $d^* \in \{32, 64\}$. The staircase at any fixed $\tau$ is a level cut of the smooth error surface $\varepsilon_{\mathrm{med}}(h, d)$ (Fig.~\ref{fig:fan}).

\begin{table}[H]
\caption{The staircase at two sizes, as $d^*(h)$ per cell at $\tau = 10^{-2}$ under the deep protocol (plotted as Fig.~\ref{fig:staircase} in the main text). The edge of the step falls inside the critical window at both sizes, and the demand relaxes to 64 deep in the paramagnet. A dash marks a column that was not run at that size. A dagger ($\dagger$) marks a value of $d^*$ decided by a cell whose median lies within $\pm 10\%$ of the threshold.}
\label{tab:staircase}
\begin{center}
\footnotesize
\setlength{\tabcolsep}{3.4pt}
\begin{tabular}{lcccccccccccccc}
\toprule
$h/J$ & 0.6 & 0.8 & 0.9 & 0.95 & 1.0 & 1.05 & 1.1 & 1.2 & 1.3 & 1.4 & 1.6 & 2.0 & 4 & 10 \\
\midrule
$L = 16$ & 32 & 64 & 64 & 64 & 64$^\dagger$ & 128 & 128$^\dagger$ & 256$^\dagger$ & 128$^\dagger$ & 128 & 128 & 128 & 64$^\dagger$ & 64 \\
$L = 24$ & 32 & --- & 64 & --- & 64 & 64$^\dagger$ & 128 & 128 & 128 & 128 & --- & --- & --- & --- \\
\bottomrule
\end{tabular}
\end{center}
\end{table}

\section{Further related work}
\label{app:related}

Other work counts how many directions or parameters an NQS needs with other tools. \citet{dash2025efficiency} take the rank of the quantum geometric tensor at convergence as the number of directions a trained NQS uses, mainly for networks trained by infidelity minimization on a spin-1 chain. \citet{park2020geometry} find that, under SR, the converged spectrum of the quantum Fisher matrix of complex restricted Boltzmann machine states can change sharply across a phase transition. \citet{barton2026connectivity} prune NQS of the transverse-field Ising model and the toric code by more than an order of magnitude and find that the connectivity of the sparse subnetwork, not its initialization, sets its accuracy. Their subnetworks are selected by pruning and set the removed weights to zero, while our slice is dense and drawn at random.

On the sign structure, \citet{szabo2020neural} give an NQS an explicit phase ansatz and find, in frustrated antiferromagnets, low-energy states that obey the Marshall sign rule and likely keep VMC from reaching the ground state. \citet{ou2025improving} train separate amplitude and phase networks on the $J_1$--$J_2$ model, with a larger SR time step for the phase. \citet{choo2019two} benchmark a convolutional NQS on the same model, including the $6\times6$ lattice.

\end{document}